\documentclass[letterpaper, 10 pt, conference]{ieeeconf}  

\IEEEoverridecommandlockouts                              

\usepackage{graphicx}
\usepackage{mathrsfs}
\usepackage{amsfonts}
\usepackage{multicol}
\usepackage{standalone}
\usepackage{xcolor}
\usepackage{eufrak}
\usepackage{amsmath} 
\usepackage{tikz}
\usepackage{pgfplots}
\usepackage{pgfplotstable}
\usepackage{leftindex}
\usepackage{multirow}
\pgfplotsset{compat=1.7}
\usepackage{subcaption}
\usepgfplotslibrary{groupplots}
\usepackage[textsize=tiny]{todonotes}

\DeclareMathAlphabet{\mathpzc}{OT1}{pzc}{m}{it}
\newcolumntype{P}[1]{>{\centering\arraybackslash}p{#1}}

\title{\LARGE \bf
Leveraging Neural Radiance Field in Descriptor Synthesis for Keypoints Scene Coordinate Regression
}

\author{Bui Huy Hoang$^{1}$, Bui Bach Thuan$^{1}$, Tran Dinh Tuan$^{2}$, Joo-Ho Lee$^{2}$
\thanks{$^{1}$Bui Huy Hoang and Bui Bach Thuan are with Graduate School of Information Science and Engineering,
        Ritsumeikan University, Japan
        }%
\thanks{$^{2}$Tran Dinh Tuan and Joo-Ho Lee are with College of Information Science, Ritsumeikan University,
        Japan
        }%
}

\begin{document}

\maketitle
\thispagestyle{empty}
\pagestyle{empty}

\begin{abstract}

Classical structural-based visual localization methods offer high accuracy but face trade-offs in terms of storage, speed, and privacy. A recent innovation, keypoint scene coordinate regression (KSCR) named D2S addresses these issues by leveraging graph attention networks to enhance keypoint relationships and predict their 3D coordinates using a simple multilayer perceptron (MLP). Camera pose is then determined via PnP+RANSAC, using established 2D-3D correspondences. While KSCR achieves competitive results, rivaling state-of-the-art image-retrieval methods like HLoc across multiple benchmarks, its performance is hindered when data samples are limited due to the deep learning model's reliance on extensive data. This paper proposes a solution to this challenge by introducing a pipeline for keypoint descriptor synthesis using Neural Radiance Field (NeRF). By generating novel poses and feeding them into a trained NeRF model to create new views, our approach enhances the KSCR's generalization capabilities in data-scarce environments. The proposed system could significantly improve localization accuracy by up to 50\% and cost only a fraction of time for data synthesis. Furthermore, its modular design allows for the integration of multiple NeRFs, offering a versatile and efficient solution for visual localization. The implementation is publicly available at: https://github.com/ais-lab/DescriptorSynthesis4Feat2Map.

\end{abstract}

\section{INTRODUCTION}


Visual (re)localization is the fundamental component within many robotics, computer vision, and augmented reality applications. Given a query image, the system outputs the 6-Degree of Freedom (DoF) of the camera with respect to the scene model. This problem has been extensively studied, and many methods have been proposed. Generally, the approaches in visual localization can be categorized into classical structure-based methods, learning-based methods, and scene coordinate regression (SCR). With 
classical structure-based approaches, given the model of scene representation constructed by Structure from Motion (SfM)\cite{schonberger2016structure}, a query image is first extracted to get the feature points. Then the feature points are matched with their corresponding 3D coordinate inside the scene model. Then, camera pose can be obtained through geometry like RANSAC-based optimization. Such methods could achieve localization results with high accuracy in multiple benchmarks. However, they often need to store the 3D scene model along with the feature descriptor which would increase storage and slow down the performance as the scene scales. 

\begin{figure}
    \centering
    \includegraphics[scale=0.7]{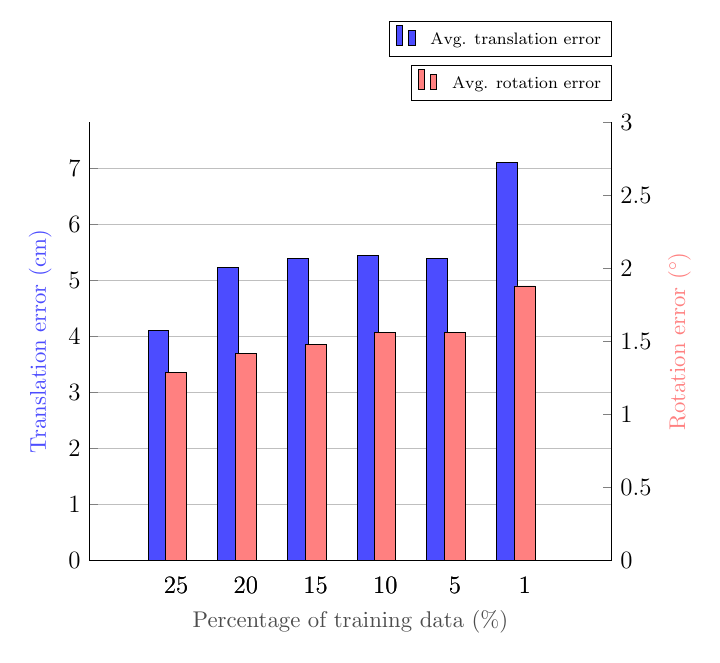}
    \caption{\textbf{Performance degradation of D2S \cite{bui_d2s_2023} when training data is reduced}. Average translation error (cm) and rotation error (degree) for all scenes in 7Scenes datasets are reported. The amount of training data ranges from 25 \% to 1\%.}
    \label{d2s_degrade}
\end{figure}

With recent advancements in deep learning, much attention in visual localization was shifted toward learning-based \cite{kendall_posenet_2016, bach_featloc_2022, zhou_kfnet_2020, zhou2020learn} and SCR approaches \cite{brachmann_dsac_2018, brachmann_learning_2018, bui_d2s_2023, li_hierarchical_2020, dong_visual_2022}. While the former addresses storage and speed concerns, they exhibit scene-specific limitations and may lag behind in terms of accuracy. In contrast, SCR approaches resolve accuracy issues in learning-based methods by initially predicting the 3D coordinates of 2D image points before proceeding to pose estimation. Notably, Bui et al. \cite{bui_d2s_2023} introduced D2S, a method that employs a model comprising a Graph Neural Network (GNN) and a Multilayer Perceptron (MLP). The GNN refines and reinforces the connections between keypoints within a frame, and the MLP is utilized to regress the coordinates. The predicted 2D-3D correspondences are then leveraged for camera pose estimation using PnP+RANSAC \cite{kukelova2013real}.

Even though D2S performs well on the majority of benchmarks, trailing slightly behind state-of-the-art structure-based approaches such as HLoc \cite{sarlin_coarse_2019}, it struggles with the acquisition of substantial training data. Similar to many deep learning architectures, D2S demands a considerable volume of data for effective generalization. The process of mapping keypoint descriptors to scene coordinates necessitates the collection of numerous images from various angles to adequately capture descriptor changes relative to different viewpoints. The graph in Fig. \ref{d2s_degrade} illustrates how D2S's performance degrades with a reduction in training data.

The introduction of Neural Radiance Fields (NeRF), as proposed in \cite{mildenhall_nerf_2020}, has significantly influenced the domains of implicit scene representation. NeRF has demonstrated its capability to generate high-quality images conditioned by camera poses, showcasing an immense potential for novel view synthesis within visual localization research. Prior efforts, such as \cite{moreau_lens_nodate}, suggested employing NeRF to render new images, leading to improvement in direct pose regression methods. However, for KSCR there has been limited exploration into leveraging NeRF to enhance performance, particularly in scenarios with limited data. Consequently, we recognized an opportunity to bridge this gap and establish a connection between NeRF and KSCR approaches.

To address the aforementioned issue, we proposed a descriptor synthesis pipeline for KSCR, specifically focusing on D2S using NeRF in scarce data scenarios. Using NeRF models such as Nerfacto \cite{tancik_nerfstudio_2023} allows the pipeline to capture the scene representation using a few images at a rapid rate. To render novel views, we propose using spherical linear interpolation to obtain new poses from camera poes in the training dataset. Instead of rerunning structure from motion (SfM), we utilized the Superpoint extractor \cite{detone_superpoint_2018} and LightGlue matcher \cite{lindenberger_lightglue_2023} to extract keypoint descriptors from novel views. This enriches the dataset and improves D2S results after training with both original and synthetic data. Our main contributions are:
\begin{itemize}
    \item We introduce a data synthesis pipeline that significantly reduces the need for collecting training data while still improving localization performance for KSCR.
    \item Utilizing NeRF and feature matcher simplifies and accelerates both view synthesis and view quality evaluation without introducing complexities to the rendering model.
    \item Extensive experiments on 2 popular indoor localization datasets show the combination of the proposed pipeline and KSCR outperforms existing SCR and few-shot approaches.
\end{itemize}



\section{RELATED WORKS}
\subsection{Classical Structure-based Visual Localization}
Classical structure-based methods typically rely on a combination of computer vision and multiview geometry techniques. Hloc \cite{sarlin_coarse_2019} is a fully developed system that follows this scheme which consists of scene modeling \cite{schonberger2016structure, schonberger2016pixelwise}, feature extraction \cite{detone_superpoint_2018, tyszkiewicz_disk_2020, dusmanu_d2-net_2019, revaud_r2d2_2019}, image retrieval \cite{arandjelovic_netvlad_2016, gordo_end--end_2017}, feature matching \cite{sarlin_superglue_2020, lindenberger_lightglue_2023} and pose estimation. Due to its modular design, different components could be replaced and improved upon. Components like feature extractor and feature matcher were dominated by deep learning approaches. The combination of the classical framework with deep learning-based components makes it a robust visual localization scheme across multiple benchmarks. However, such methods raise privacy concerns and also storage when dealing with large-scale scenes. Attempts to address the memory issue can be found in \cite{bergamo2013leveraging}. GoMatch \cite{brahmbhatt_geometry-aware_2018} succeeds in solving privacy problems by only using geometric information but accuracy still falls behind by a large margin.

\subsection{Learning-based Visual Localization}
Learning-based approaches replace the need for storing and retrieving information from the scene model by encoding it to implicit representation. With direct pose estimation, PoseNet \cite{kendall_posenet_2016} was the first to propose using CNN to map the input query image to the camera's absolute pose. Other improvements can be found in \cite{bach_featloc_2022, wang2020atloc}. Although the direct approach is more memory efficient and fast compared to classical structure-based methods, their properties make them fall behind in terms of accuracy as pointed out in \cite{sattler_understanding_2019, ng_reassessing_2021}.

\subsection{Scene Coordinate Regression}

SCR addresses the accuracy challenges of previous learning-based methods by training a model to predict the scene coordinate of pixels instead of absolute pose. DSAC \cite{brachmann_dsac_2018, brachmann2021dsacstar}, was the first to offer a differential RANSAC pipeline for dense SCR. Another novel approach in SCR was presented by Bui et al. \cite{bui_d2s_2023}, suggesting that salient and robust key points suffice for visual localization. This method achieves comparable results with state-of-the-art classical structure-based approaches in most benchmarks but faces challenges in environments with similar textures or patterns. Additional research explores SCR using hierarchical scene representations \cite{li_hierarchical_2020} or employing an agnostic feature extractor \cite{dong_visual_2022}.

\subsection{Data Synthesis for Visual Localization}
\begin{figure*}[thpb]
  \centering
  \includegraphics[scale=0.65]{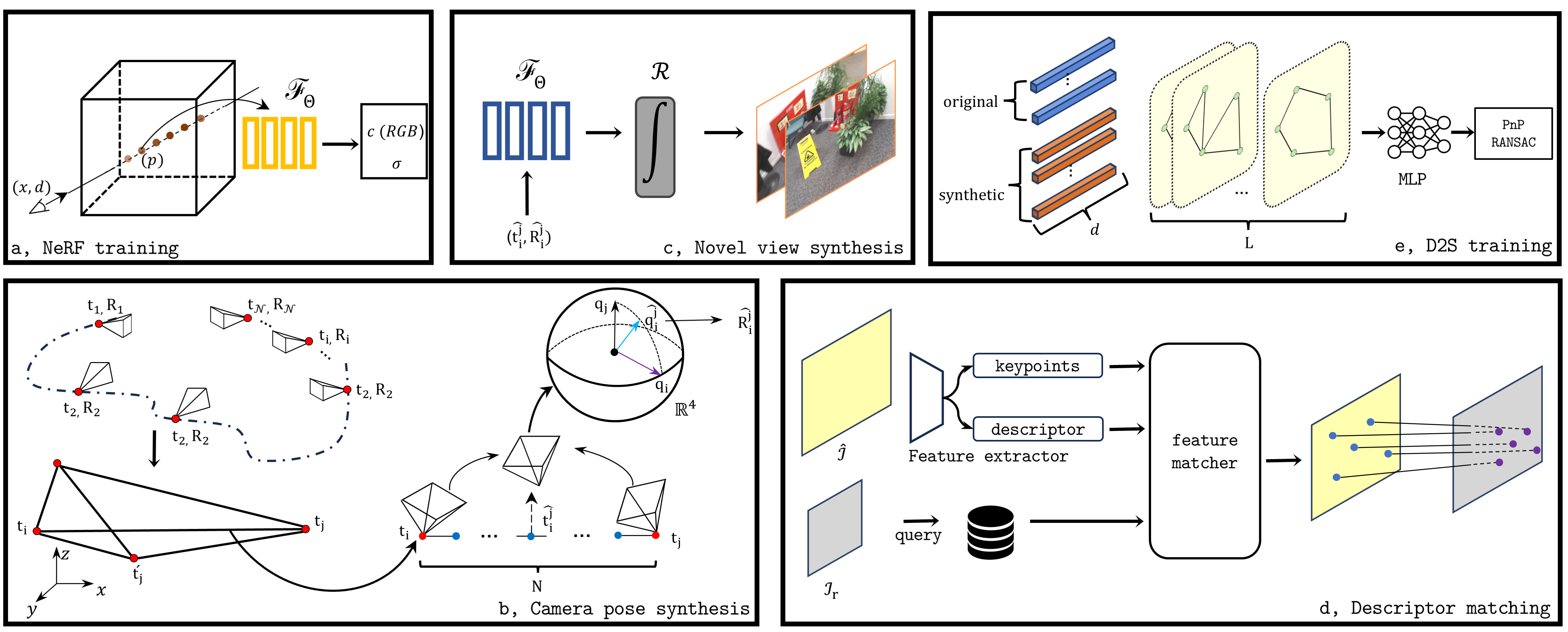}
  \caption{\textbf{Descriptor synthesis pipeline for D2S}. First, the NeRF model is trained with available images in the dataset (a). Then new camera poses are generated using a uniform sample between translation and quaternion interpolation between poses (b). Next, new poses are fed into the trained NeRF for novel view synthesis (c). 2D-2D correspondences between reference frames and novel frames are established and descriptors from novel frames are extracted (d). Finally, D2S is trained with original data and synthesized data (e).}
  \label{descriptor_synthesis_pipeline}
\end{figure*}

Traditional data augmentation in computer vision falls short in visual localization due to the disparity between transformations in 2D images and the 3D world. An U-Net-like model was proposed in \cite{pittaluga_revealing_2019} to reconstruct the image based on the reprojected color and descriptors. The utilized model was heavy and slow to train due to multiple U-Net being concatenated for detect visibility and coarse to fine reconstruction. Zhang et al. \cite{zhang2022rendernet} utilized nearby views for augmentation, distilling features to generate similar descriptors as original images. The usage of Generative Adversarial Networks is also explored in \cite{liu_posegan_2020}. Recent work with NeRF has also made contributions to visual localization as in \cite{moreau_lens_nodate, chen_leveraging_2023}.

\subsection{Visual Localization with Scarce Data}
The ability to learn using only a small amount of data in learning-based visual localization received much attention recently because it greatly reduced the effort for collecting and labeling data for the training process. Besides methods to enrich the available dataset using data augmentation \cite{bui_d2s_2023}, data synthesis \cite{moreau_lens_nodate, liu_posegan_2020, pittaluga_revealing_2019}, few-shot learning is also a potential approach. Recent work \cite{dong_visual_2022} used hierarchical partition trees which treat the scene coordinate prediction as classification rather than regression.

\section{METHODOLOGY}

This section outlines our descriptor synthesis pipeline from a limited number of images. A detailed illustration of the system is provided in Fig. \ref{descriptor_synthesis_pipeline}.


\subsection{NeRF Training}
NeRF model is utilized to learn the implicit scene representation. Given that we have a set of of images $\mathcal{I}=\{\mathcal{I}_1, \mathcal{I}_2, ..., \mathcal{I_{\mathcal{N}}}\}$ where $\mathcal{N}$ is the number of spare views in the dataset. Using SfM, we can deduce the camera parameters set $\mathbf{C}=\{\mathbf{C_1, \mathbf{C_2}, ..., \mathbf{C_{\mathcal{N}}}}\}$ where $\mathbf{C}_i$ consists of intrinsic parameter $\mathbf{K}_i\in\mathbb{R}^{3\times3}$ and extrinsic parameter $\mathbf{T}_i\in\mathbb{R}^{4\times4}$. The extrinsic parameter is composed of translation $t$ and rotation component $\mathbf{R}$ as follows:
\begin{equation}
\mathbf{T_i}=\begin{bmatrix}\mathbf{R_i}& t_i\\ 0^T & 1\end{bmatrix}
\end{equation}
Conventional NeRF utilized a continuous function to model the scene as:
\begin{equation}
   \mathscr{F}_{\Theta}:(x, d)\rightarrow(c, \sigma) 
\end{equation}
where $x$ and $d$ are the translation and view angle. $c$ represents color $(r, g, b)$ while $\sigma$ represents the density.
In order to render a pixel of an image $\hat{\mathcal{I}}$, NeRF utilizes the rendering function $\mathcal{R}$. It approximates the integral by accumulating the radiance and density of sampled points along the ray shoot from camera position $o$, passing through pixel $p$. Then, NeRF can be trained by minimizing the photometric loss.
\begin{equation}
\mathcal{L}=\sum_i^{\mathcal{N}}(||\mathcal{I}_i-\hat{\mathcal{I}}_i||_2^2)
\end{equation}

The vanilla NeRF model could render novel views within bounded scenes but encounters challenges in unbounded scenes, leading to unwanted artifacts. Moreover, it demands extensive training time. Approaches like mip-NeRF \cite{barron_mip-nerf_2021, barron_mip-nerf_2022} aim to mitigate artifact issues, while Instant-NGP \cite{muller_instant_2022} address computational overhead, reducing training time to minutes. Leveraging advancements in neural rendering research, we opt for the Nerfacto model provided by Nerfstudio \cite{tancik_nerfstudio_2023}, which integrates these improvements into the NeRF framework.

\subsection{Camera Pose Synthesis}
To generate novel views, a novel camera pose is required. Using the camera parameter set $\mathbf{C}$, our objective is to create new viewpoints. The synthesis process is depicted in Fig. \ref{descriptor_synthesis_pipeline}b. The extrinsic parameter is first decomposed into a set of reference translation $\{\mathbf{t}_i\in \mathbb{R}^3\}^{\mathcal{N}}$ and rotation $\{\mathbf{R}_i\in \mathbb{R}^{3\times3}\}^{\mathcal{N}}$. For every reference translation vector, we calculate the distance of it and others in $\{\mathbf{t}_i\}^{\mathcal{N}}$.
\begin{equation}
\mathcal{D}(\mathbf{t}_i)=\{\mathbf{d}(\mathbf{t}_i, \mathbf{t}_j); 0\leq j\leq \mathcal{N}-1; \mathbf{t}_i\neq \mathbf{t}_j\}
\end{equation}
where $\mathbf{d}(\mathbf{t}_i, \mathbf{t}_j)=||\mathbf{t}_i-\mathbf{t}_j||^2_2$ i.e Euclidean distance between two vectors.
Then we select the top $k$ translation vector closest to $\mathbf{t}_i$ based on the calculated distance to form a list of pairs.
\begin{multline}
    \mathcal{S}(\mathbf{t}_i)=\{(\mathbf{t}_i, \mathbf{t}_j):|\{(\mathbf{t}_i, \mathbf{t}_m): \\ \mathbf{d}(\mathbf{t}_i, \mathbf{t}_m)\in \mathcal{D},\,\mathbf{d}(\mathbf{t}_i, \mathbf{t}_j)>\mathbf{d}(\mathbf{t}_i, \mathbf{t}_m)\}|<k\}
\end{multline}
After $\mathcal{S}$ is obtained, for every reference pair $(\mathbf{t}_i, \mathbf{t}_j)$ in $\mathcal{S}$ we uniform sample from the line connect between those 2 points.
\begin{equation}
    \hat{\mathbf{t}}_{i}^j=\left\{(\mathbf{t}_i+\frac{n}{\mathbf{N}+1}(\mathbf{t}_j-\mathbf{t}_i)):0 < n < {\mathbf{N}+1}; n\in \mathbb{N}^*\right\}
    \label{translation}
\end{equation}
where $\mathbf{N}$ is the number of wanted samples between 2 reference translation vectors. 
Next, we proceed to perform spherical interpolation of quaternion for every sampled vector $\hat{\mathbf{t}}_{i}^j$. We first get the rotation component of 2 reference translation vectors that construct $\hat{\mathbf{t}}_{i}^j$, we denote as $\mathbf{R_i}, \mathbf{R_j}$ and convert them to quaternion $\mathbf{q}_i, \mathbf{q}_j$, respectively. The conversion from rotation matrix $\left[r_{ij}\right]_{1\leq i, j \geq 3}$ to $\mathbf{q}(w, x, y, z)$ is done by using Eq. \ref{q_to_e}:
\begin{equation}
    \mathbf{q} = \frac{1}{||\mathbf{q}||}\dot(w, x, y, z)
    \label{q_to_e}
\end{equation}
where $w =\sqrt{1+r_{11}+r_{22}+r_{33}}$; $x = (r_{32}-r_{23})/(4w)$; $y = (r_{13}-r_{31})/(4w)$; and $z = (r_{21}-r_{12})/(4w)$.

The interpolated quaternion $\hat{\mathbf{q}}_i^j$ for $\hat{\mathbf{t}}_i^j$ is calculated using $\mathbf{q}_i$ and $\mathbf{q}_j$ as follow:
\begin{equation}
    \hat{\mathbf{q}_i^j}=\frac{\sin((1-\delta)w)\mathbf{q}_i+\sin(\delta w)\mathbf{q}_j}{\sin(w)}; \delta \in [0, 1]
    \label{quaternion}
\end{equation}
where $\delta=\left\{i/(N+1)\right\}^\mathbf{N}_{1}$.

From Eq. \ref{translation} and Eq. \ref{quaternion}, we obtained the new poses $(\hat{\mathbf{t}}_i^j, \hat{\mathbf{q}}_i^j)$ for novel view synthesis.
\subsection{Novel View Synthesis}
Using the trained NeRF model from section A and the generated viewpoints from section B, we synthesize novel views as illustrated in Fig. \ref{descriptor_synthesis_pipeline}c. The set of new views conditioned by the poses is obtained by rendering function $\mathcal{R}$.
\begin{equation}
    \hat{\mathcal{I}}_i = \mathcal{R}((\hat{\mathbf{t}}_i^j, \hat{\mathbf{q}}_i^j)|\Theta)
\end{equation}

\subsection{Descriptor Matching}
From the set of synthesized images, we extract the features by performing image matching with the available images in the database. We first consider 2 reference camera pose $(\mathbf{t}_r, \mathbf{q}_r)$ which produce sampled pose $(\hat{\mathbf{t}}, \hat{\mathbf{q}})$. Their corresponding images are $\mathcal{I}_r$, and $\hat{\mathcal{I}}$, respectively. The keypoints and corresponding descriptors of $\hat{\mathcal{I}}$ are extracted using feature extractor and denoted as $\{\hat{\mathbf{p}}_i\in[0,1]^2\}^{\mathcal{N}_F}$ and $\{\hat{\mathbf{d}}_i\in \mathbb{R}^d\}^{\mathcal{N}_F}$ where $\mathcal{N}_F$ is the number of detected features and $d$ is the descriptor dimension. At the same time, keypoints and descriptors of reference poses are retrieved from the SfM database and denoted as $\{\mathbf{p}_i\}^{\mathcal{M}_F}$, $\{\mathbf{d}_i\}^{\mathcal{M}_F}$ where $\mathcal{M}_F$ is the number of retrieved features. We seek to match the feature from image $\mathcal{I}_r$ to image $\hat{\mathcal{I}}$ and output a set of correspondence index:
$$
\mathbf{M}=\{(i, j)\subset \{{1,...,\mathcal{N}_F}\} \times\{1,...,\mathcal{M}_F\}\}
$$
During the matching process, we determine the quality of the synthetic image $\hat{\mathcal{I}}$ by setting the threshold for the number of matches $\eta$. If the number of matches is lower than $\eta$, the image is discarded. This allows us to filter out the poor quality ones that contain many artifacts or occlusion. After matching we collect the matched keypoints and descriptors of the synthesized image $\{(\hat{\mathbf{p}}_i, \hat{\mathbf{d}}_i)\}^{\mathcal{N}_{matched}}$.

\subsection{Joint Training}
After the keypoints and their corresponding descriptors are extracted from the synthesized images, we proceed to train KSCR with original and synthetic data as illustrated in Fig. \ref{descriptor_synthesis_pipeline}e. D2S model aim to learn the function $\mathfrak{F}:\mathbb{R}^{\mathcal{M}_F\times d}\rightarrow \mathbb{R}^{\mathcal{M}_F\times 4}$ that map keypoints's descriptor $\{\mathbf{d}_i\}^{\mathcal{M}_F}$ in a image to their corresponding scene coordinate $w=(x, y, z, p)^T$, where $p$ is the uncertainty of the descriptor $\mathbf{d}$.
The model learns the optimal parameter $\Theta_K$ by minimizing the L2 loss between the predicted coordinate $\hat{w}$ and ground truth $w$, uncertainty loss for descriptors, and reprojection loss for camera pose, as detailed in \cite{bui_d2s_2023}.

\section{EXPERIMENTS}
\subsection{Network Configuration}
\textbf{NeRF}: For novel view synthesis, we employ the Nerfacto model provided in the Nerfstudio framework. The model comprises of Camera Optimizer from NeRF-- \cite{wang2021nerf}, Scene Contraction and Ray Sampler from MipNeRF-360\cite{barron_mip-nerf_2022, barron_mip-nerf_2021}, Appearance Embedding from NeRF-W and HashEncoding, FusedMLP from Instant-NGP \cite{muller_instant_2022}.

\textbf{Feature Extraction and Matching}: In order to keep the consistency with the provided implementation in D2S, we use Superpoint \cite{detone_superpoint_2018} as the extractor. For keypoints matching between the reference frame and the synthesized frame, we choose LightGlue because it provides performance on par with heavier models like SuperGlue \cite{sarlin_superglue_2020} while being faster. 

\textbf{Keypoint Scene Coordinate Regressor}: We use the original implementation of D2S which consists of a GNN and MLP stacked on top. The number of layers for the GNN is 5, while the MLP is composed of 5 layers with sizes $(512, 1024, 1024, 512, 4)$. The initial learning rate of the D2S model is $5\times 10^{-4}$ and it gradually decays with the decay parameter of 0.5. The batch size is 4 and the optimizer is Adam. 

All experiments were conducted on the same machine equipped with an Intel Core i7-8700K CPU and a Nvidia RTX 3080ti GPU. 


\begin{figure*}
    \centering
    \begin{subfigure}{.18\textwidth}
          \centering
          \includegraphics[scale=0.4]{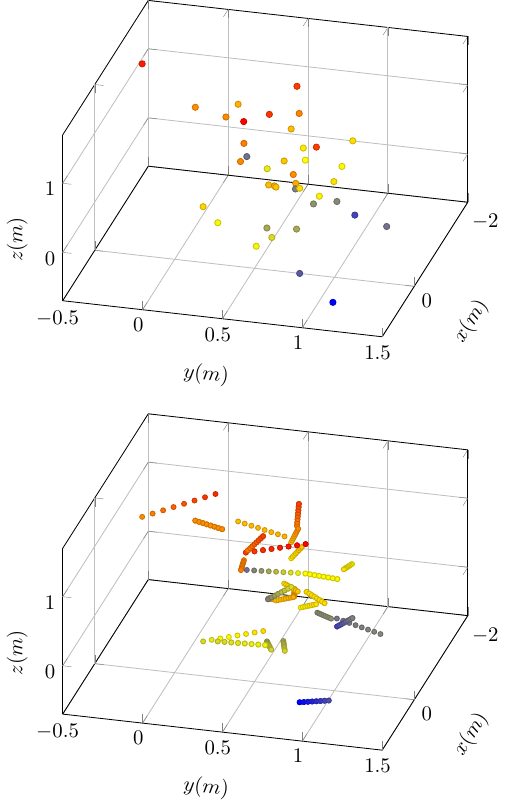}
          \caption{7S-Chess}
          \label{chess_pose}
    \end{subfigure}
    \begin{subfigure}{.18\textwidth}
          \centering
          \includegraphics[scale=0.4]{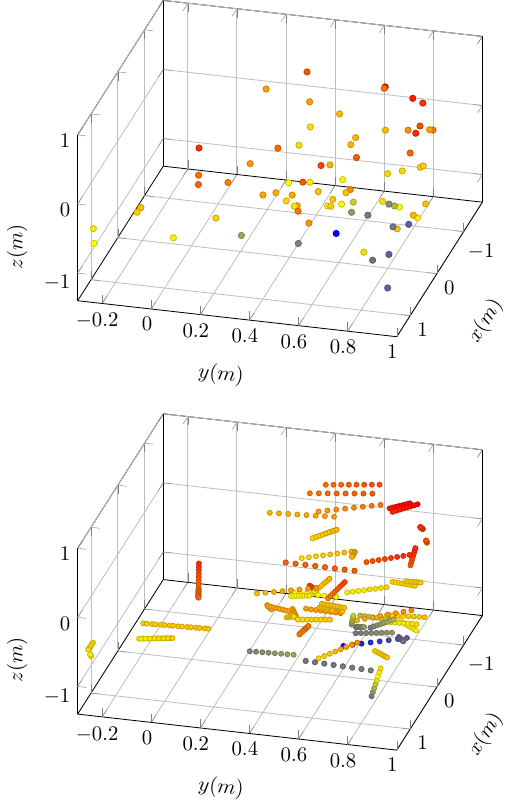}
          \caption{7S-Pumpkin}
          \label{pumpkin_pose}
    \end{subfigure}
    \begin{subfigure}{.18\textwidth}
          \centering
          \includegraphics[scale=0.4]{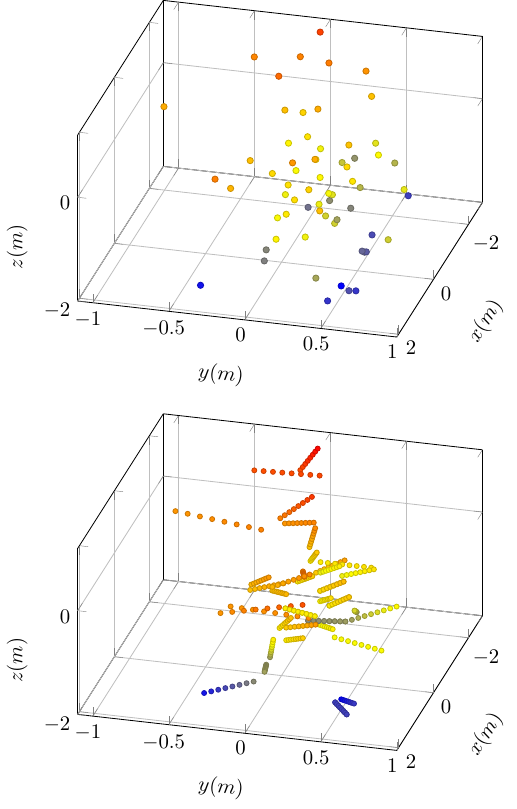}
          \caption{7S-Redkitchen}
          \label{redkitchen_pose}
    \end{subfigure}
    \begin{subfigure}{.18\textwidth}
          \centering
          \includegraphics[scale=0.4]{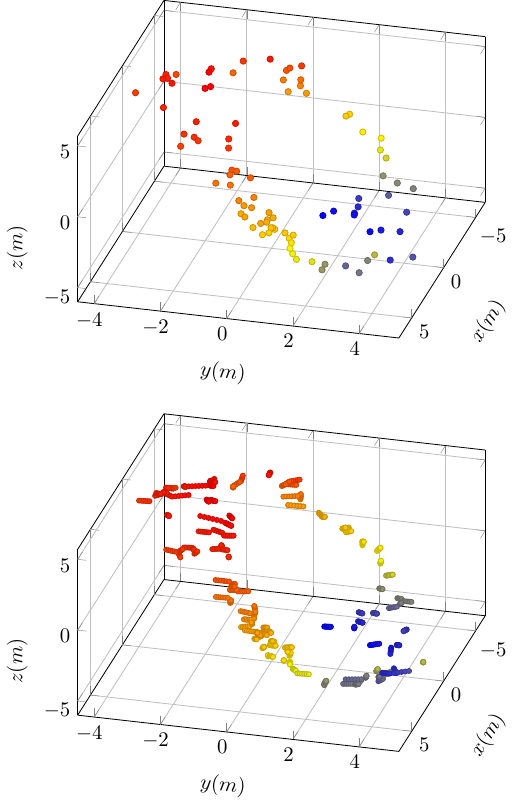}
          \caption{12S-Gates362}
          \label{office1_362_pose}
    \end{subfigure}
    \begin{subfigure}{.18\textwidth}
          \centering
          \includegraphics[scale=0.4]{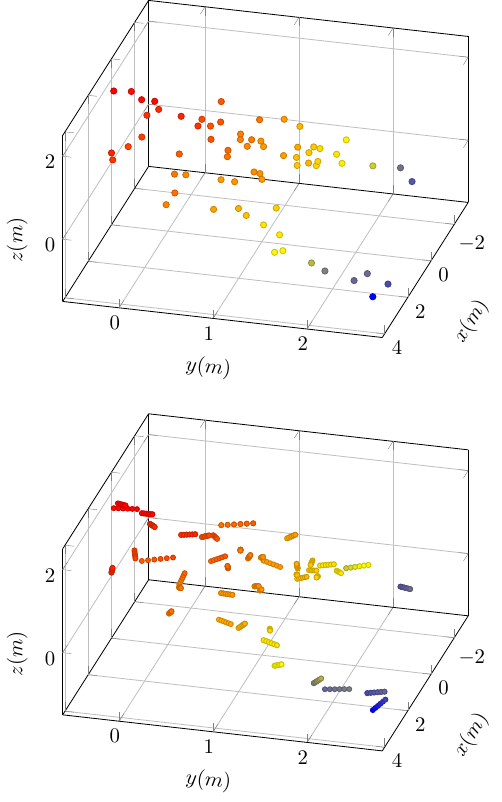}
          \caption{12S-Office5b}
          \label{office2_5b_pose}
    \end{subfigure}
    \caption{\textbf{Illustration of camera position of training images and generated camera position}. The top row depicts the camera position of training images, and the bottom row shows the generated camera position for NeRF rendering.}
    \label{camera_position}
\end{figure*}

\subsection{Dataset}
We evaluate our method using two standard indoor visual-localization datasets, 7Scenes \cite{shotton_scene_2013} and 12Scenes \cite{valentin2016learning}.

\textbf{7Scenes}: The dataset comprises 7 sets of RGB-D images obtained through KinectFusion. The recorded environment spans volumes ranging from $1m^3$ to $18m^3$. Each recording consists of a varying number of training images, ranging from $1000$ to $7000$ images. In our experiments, we intentionally operate in scenarios with limited data. Specifically, for each scene in the original dataset, we use uniform sampling method \cite{chen_leveraging_2023} to sample $1\% \sim 2\%$ of the original data.

\textbf{12Scenes}: Similar to 7Scenes, this dataset comprises RGB-D images, but it is characterized by more diversity and larger volumes. In a similar manner, we uniformly sample $5\%$ of each scene from the original data for training.

\subsection{Data Synthesis Settings}

During benchmarking on the 7Scenes dataset, we maintain consistency in the hyperparameters of the descriptor synthesis pipeline. Specifically, we set top k translation pairs to 3 and the number of samples between 2 points $\mathbf{N}=40$. The matching threshold $\eta$ is fixed at $500$. For descriptor matching, we align the number of maximum extracted keypoints of Superpoint with the configuration in \cite{bui_d2s_2023}, which is 2048. Additionally, network pruning, aimed at accelerating the matching process of LightGlue \cite{lindenberger_lightglue_2023}, is disabled to ensure optimal matching results.

\begin{table}[h]
\centering
\caption{Number of images as training, synthetic, and after filtering in 7Scenes and 12Scenes dataset}
\label{table_number_of_images}
\begin{tabular}{|m{0.5cm}|m{1.2cm}|P{1.2cm}|P{1.2cm}|P{1.2cm}|}
\hline
 & Scenes & \# training images & \# synthesized images & \# images after filtering\\
\hline
\multirow{7}{*}{7S} & {Chess} & {40}& {1600}& {1476} \\
                    & {Fire} & {20} & {800} & {688} \\
                    & {Heads} & {20} & {800} & {424} \\
                    & {Office} & {60} & {2400} & {1530} \\
                    & {Pumpkin} & {80} & {3200} & {2710} \\
                    & {Redkitchen} & {70} & {2720} & {1279} \\
                    & {Stairs} & {40} & {1600} & {1179}\\\hline

\multirow{12}{*}{12S} & {Kitchen1} & {37}& {1110}& {816} \\
                    & {Living1} & {51} & {1530} & {1480} \\
                    & {Bed} & {44} & {1320} & {1046} \\
                    & {Kitchen2} & {39} & {1170} & {892} \\
                    & {Living2} & {36} & {1080} & {770} \\
                    & {Luke} & {68} & {2040} & {1615} \\
                    & {Gates362} & {177} & {5310} & {4876} \\
                    & {Gates381} & {147} & {4410} & {3563} \\
                    & {Lounge} & {46} & {1380} & {1020} \\
                    & {Manolis} & {81} & {2430} & {1748} \\
                    & {Office5a} & {50} & {1500} & {815} \\
                    & {Office5b} & {68} & {2040} & {1639}\\\hline
\end{tabular}
\end{table}

\subsection{Localization Baselines}

We assess our method by comparing it with the original D2S implementation \cite{bui_d2s_2023}. Additionally, we evaluate SCRNet and HSCNet \cite{li_hierarchical_2020} as they represent SCR approaches. SRC \cite{dong_visual_2022} is considered due to its notable few-shot performance comparable to Hloc \cite{sarlin_coarse_2019}. The assessment of SRC \cite{dong_visual_2022} on 12Scenes is omitted because they used the dataset to pre-train the classification network with other datasets for model initialization. We also include a comparison with a method utilizing NeRF for view synthesis as presented in \cite{chen_leveraging_2023}. However, our comparison is constrained to different dataset settings, as the method's code is not publicly accessible.

\subsection{Data Synthesis Result}
In this section, we present the result of the descriptor synthesized pipeline. Table \ref{table_number_of_images} lists the number of images for training, images that are synthesized by NeRF, and the final count after filtering out failed cases for both the 7Scenes and 12Scenes datasets. 

The number of training images also corresponds to the number of camera poses utilized for generating new viewpoints. The training camera and generated camera position for Nerfacto can be observed in Fig. \ref{camera_position}. Notably, from a sparse set of camera positions in a scene, our camera pose synthesis effectively connects these positions and adds more viewpoints to enhance scene coverage. In Fig. \ref{camera_position}, the camera pose positions for the depicted scene in the 12Scenes dataset are denser and closer to each other, reflecting the higher number of training camera poses compared to the 7Scenes dataset.

\begin{figure}
\centering
\begin{subfigure}{.15\textwidth}
  \centering
  \includegraphics[scale=0.12]{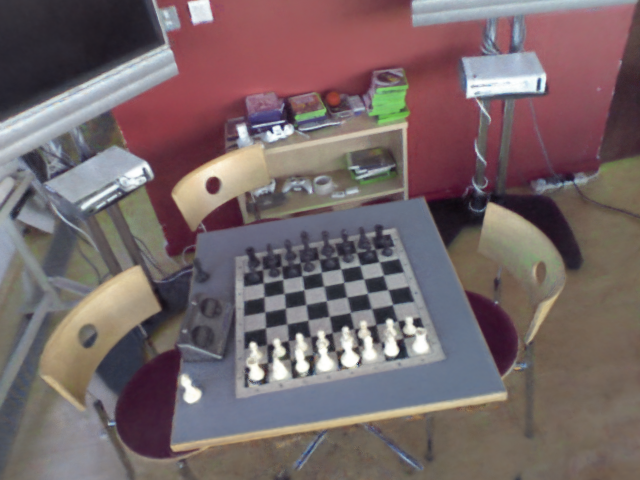}
  \caption{}
  \label{chess_good}
\end{subfigure}
\begin{subfigure}{.15\textwidth}
  \centering
  \includegraphics[scale=0.12]{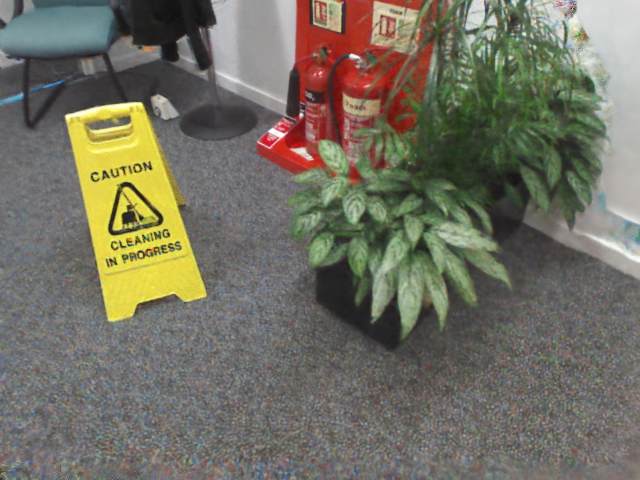}
  \caption{}
  \label{fire_good}
\end{subfigure}
\begin{subfigure}{.15\textwidth}
  \centering
  \includegraphics[scale=0.12]{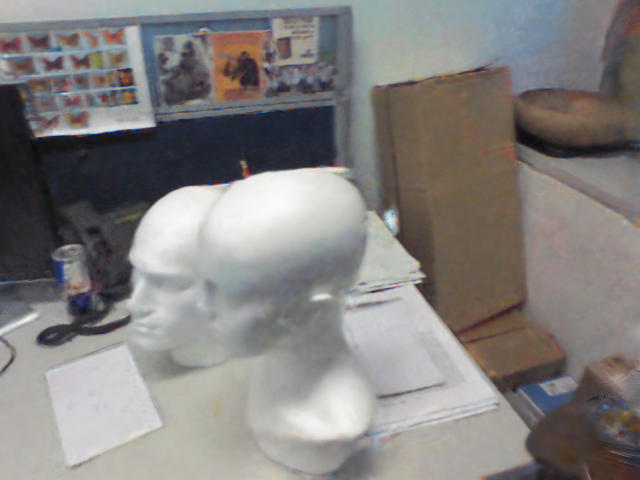}
  \caption{}
  \label{heads_good}
\end{subfigure}
\begin{subfigure}{.15\textwidth}
  \centering
  \includegraphics[scale=0.12]{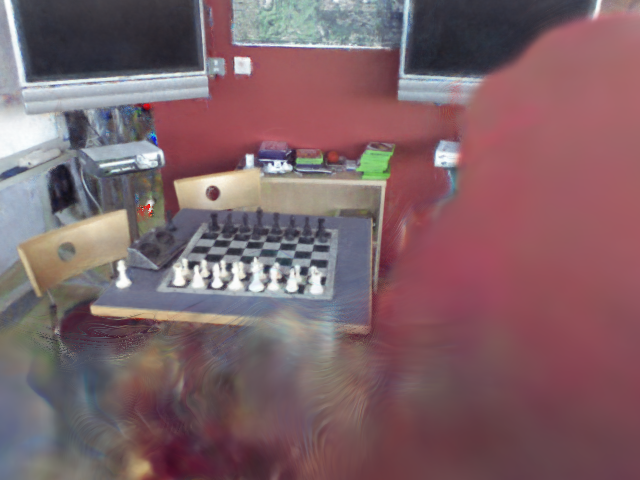}
  \caption{}
  \label{chess_fail}
\end{subfigure}
\begin{subfigure}{.15\textwidth}
  \centering
  \includegraphics[scale=0.12]{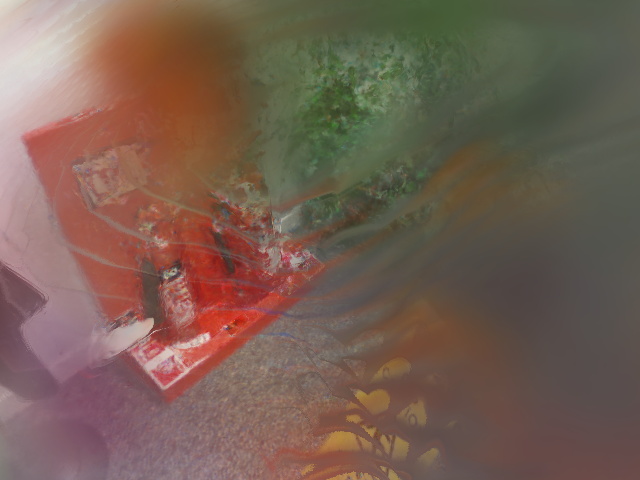}
  \caption{}
  \label{fire_fail}
\end{subfigure}
\begin{subfigure}{.15\textwidth}
  \centering
  \includegraphics[scale=0.12]{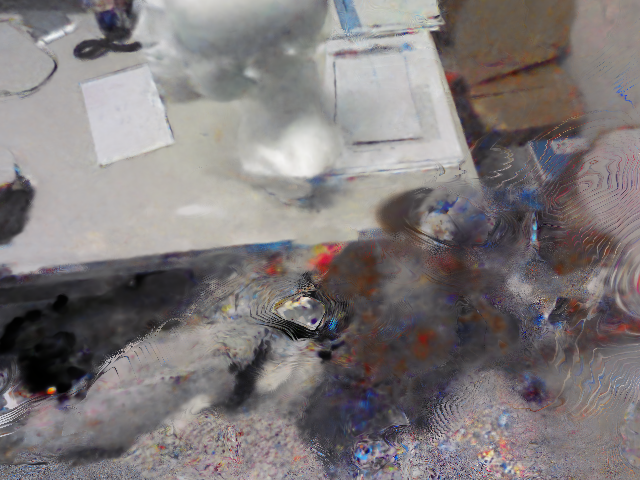}
  \caption{}
  \label{heads_fail}
\end{subfigure}
\caption{\textbf{Synthesized images from NeRF}. The columns from left to right illustrate synthesized images for Chess, Fire, and Heads respectively. The first row shows favorable results while the second presents filtered fail cases.}
\label{success_vs_fail_synthesis}
\end{figure}

Using the synthesized camera poses, we present the NeRF rendering results in Fig. \ref{success_vs_fail_synthesis}. By visual inspection, it can be seen that the images from the top row exhibit a realistic appearance that closely resembles the actual scene. However, some minor artifacts are still visible in the rendered images. Despite incorporating advanced features in Nerfacto, artifacts remain challenging, particularly due to the limited number of training images and the presence of motion blur in some instances. 

\begin{figure}
    \centering
    \begin{subfigure}{0.15\textwidth}
    \centering
        \makebox[\linewidth]{\includegraphics[scale=0.35]{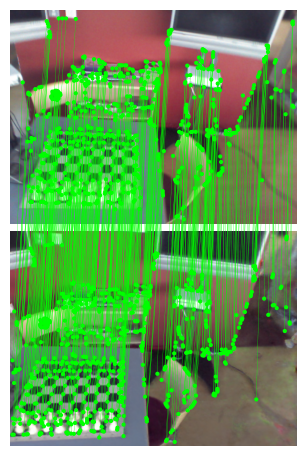}}
        \caption{}
        \label{chess_matches}
    \end{subfigure}
    \begin{subfigure}{0.15\textwidth}
        \makebox[\linewidth]{\includegraphics[scale=0.35]{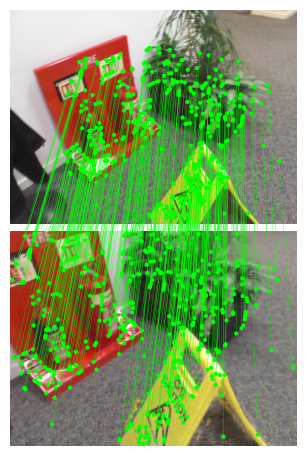}}
        \caption{}
        \label{fire_matches}
    \end{subfigure}
    \begin{subfigure}{0.15\textwidth}
        \makebox[\linewidth]{\includegraphics[scale=0.35]{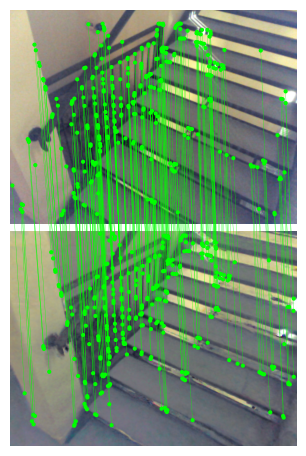}}
        \caption{}
        \label{stairs_matches}
    \end{subfigure}
    \caption{\textbf{Keypoints matching result between real and synthetic images}. The top row illustrates reference images while the bottom row presents synthesized images.}
    \label{matching_result}
\end{figure}

The bottom row of Fig. \ref{success_vs_fail_synthesis} illustrates cases where Nerf fails. Synthesized images that are obstructed or contain significant artifacts are filtered out based on the number of matches with the reference images. Unlike the approach in \cite{chen_leveraging_2023} that needs an additional head for NeRF to predict the uncertainty in order to select high-fidelity synthesized views, our method employs existing feature matcher which is more straightforward and simplifies the generated view quality evaluation. As can be seen in Table \ref{table_number_of_images}, the number of images after filtering reduced quite drastically for scenes like 7S-Office, 7S-Redkitchen, and 12S-Office5a, which have larger volumes compared to other scenes.

\begin{figure}
    \centering
    \begin{subfigure}{0.25\textwidth}
    \centering
        \makebox[\linewidth]{\includegraphics[scale=0.7]{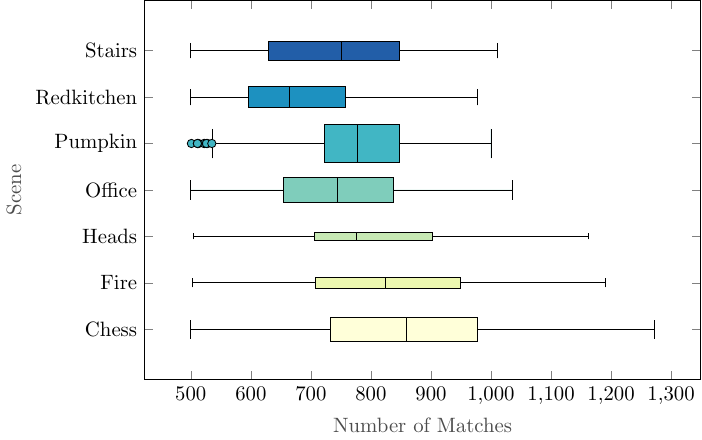}}
        \caption{7Scenes}
        \label{chess_matches}
    \end{subfigure}
    \begin{subfigure}{0.25\textwidth}
    \centering
          \makebox[\linewidth]{\includegraphics[scale=0.7]{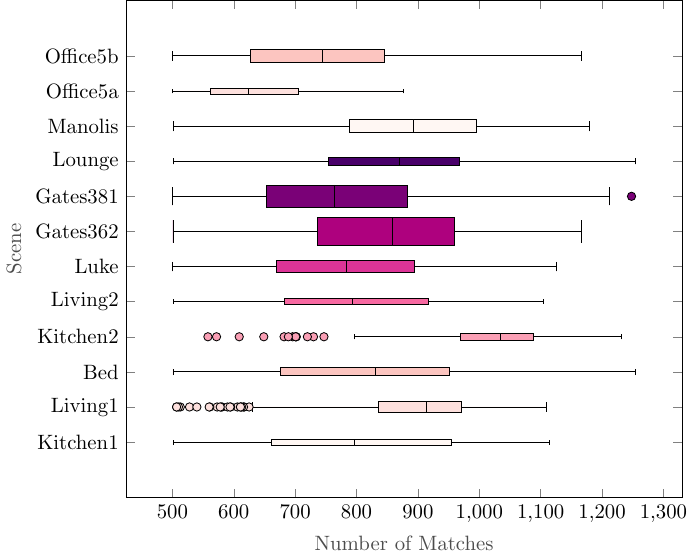}}
        \caption{12Scenes}
        \label{chess_matches}
    \end{subfigure}
    \caption{\textbf{Matched keypoints distribution of synthetic images across scene}. The size of the boxplot corresponds to the number of synthetic images.}
    \label{box_plot}
\end{figure}


Figure \ref{matching_result} illustrates examples of matched keypoints between the reference images and the synthesized images. As the generated camera poses lie between the poses of two reference images, a significant amount of similar visual content is shared, resulting in a high number of matched keypoints. However, it is crucial to note that minor artifacts could lead to incorrect matches, thus filter is used to ensure image quality.

\begin{table*}[t]
  \centering
    \caption{\textbf{Localization results on 7Scenes}. The median translation and rotation error in $\mathbf{cm}$ and $\mathbf{degrees}$ are reported. The result in \textcolor{red}{red} and \textcolor{blue}{blue} indicate the best and second best results of learning-based approaches using the same data settings.}
  \begin{tabular}{|m{1.5cm}|P{2cm}|P{2cm}|P{2cm}|P{2cm}|P{2cm}|P{2cm}|}
    \hline
    \multicolumn{1}{|l|}{} & \multicolumn{1}{P{2cm}|}{25\% training images} & \multicolumn{2}{l}{} & \multicolumn{3}{l|}{1$\sim$2\% training images}\\
    \hline
     Scene & E-SCRNet + U-NeRF \cite{chen_leveraging_2023} &  SRC \cite{dong_visual_2022} & HSC-Net \cite{li_hierarchical_2020} & SCR-Net \cite{li_hierarchical_2020} & D2S\textsuperscript{original} \cite{bui_d2s_2023} & \textbf{D2S+Descriptor Synthesis\textsuperscript{our}} \\
    \hline
    Chess & $2.7/0.9$ & $3.26/1.09$ & $2.99/0.95$ & $4.20/1.32$ & \textcolor{blue}{$2.72/0.91$} & {\textcolor{red}{$2.29/0.85$}} \\
    Fire & $2.6/1.1$ & \textcolor{blue}{$3.28/1.2$} & $3.54/1.39$ & $15.28/4.39$ & $3.40/1.33$ & {\textcolor{red}{$2.69/1.03$}} \\
    Heads & $1.3/1.0$ & $1.80/1.33$ & $3.10/2.2$ & $48.67/27.02$ & \textcolor{blue}{$1.68/1.04$} & \textcolor{red}{$1.41/0.96$} \\
    Office & $3.8/1.1$ & \textcolor{red}{$4.19/1.22$} & $5.39/1.43$ & $133.5/29.34$ & $5.05/1.39$ & \textcolor{blue}{$4.67/1.3$} \\
    Pumpkin & $5.2/1.3$ & \textcolor{blue}{$4.73/1.26$} & $6.23/1.67$ & $22.55/5.16$ & $5.74/1.48$ & \textcolor{red}{$4.66/1.19$} \\
    Redkitchen & $6.8/1.9$ & \textcolor{red}{$5.76/1.65$} & \textcolor{blue}{$6.23/$}$1.697$ & $70.10/13.07$ & $7.02/1.79$ & $6.40$/\textcolor{blue}{$1.68$} \\
    Stairs & $4.5/1.1$ & \textcolor{red}{$5.07/1.38$} & $26.37/6.69$ & $75.10/19.26$ & $24.19/5.23$ & \textcolor{blue}{$19.75/4.85$} \\
    \hline
  \end{tabular}
  \label{comparison}
\end{table*}

We show the distribution of keypoint matches between synthetic and reference images in Fig. \ref{box_plot}. Notably, the lower whisker for most scenes is situated at 500, aligning with the threshold we established. However, exceptions are evident in specific scenes such as 7S-Pumpkin, 12S-Kitchen2, and 12S-Living1. While the median value of scenes is usually in the range from 700 to 800, examining closely reveals a trend where scenes with larger volumes tend to have slightly lower median values compared to smaller volume scenes. This observation has effects on localization results, as a larger environment causes a lower number of matches, resulting in a comparatively smaller improvement.

\begin{table*}[t]
  \centering
    \caption{\textbf{Localization results on 12Scenes}. The median translation and rotation error in $\mathbf{cm}$ and $\mathbf{degrees}$ are reported. The result in \textcolor{red}{red} and \textcolor{blue}{blue} indicate the best and second best results of learning-based approaches using the same data settings. SRC \cite{dong_visual_2022} is omitted due to the usage of the 12Scenes dataset for pre-train.} 
  \begin{tabular}{|m{1.5cm}|P{2cm}|P{1.5cm}|P{2cm}|P{2cm}|P{2cm}|P{2cm}|}
    \hline
    \multicolumn{1}{|l|}{} & \multicolumn{1}{P{2cm}|}{5$\sim$20\% training images} & \multicolumn{2}{l}{} & \multicolumn{3}{l|}{5\% training images}\\
    \hline
     Scene & E-SCRNet+U-NeRF \cite{chen_leveraging_2023} &  SRC \cite{dong_visual_2022} & HSC-Net \cite{li_hierarchical_2020} & SCR-Net \cite{li_hierarchical_2020} & D2S\textsuperscript{original} \cite{bui_d2s_2023} & \textbf{D2S+Descriptor Synthesis\textsuperscript{our}} \\
    \hline
    Kitchen1 & $0.9/0.5$ & $-$ & \textcolor{blue}{$1.75$}$/0.89$ & $13.3/6.64$ & \textcolor{blue}{$1.75/0.80$} & \textcolor{red}{$0.88/0.39$} \\
    Living1 & $2.1/0.6$ & $-$ & $1.93/0.65$ & $12.9/4.33$ & \textcolor{blue}{$1.58/0.54$} & \textcolor{red}{$0.93/0.32$} \\
    Bed & $1.6/0.7$ & $-$ & \textcolor{blue}{$2.26/0.99$} & $20.50/9.17$ & $2.32/0.98$ & \textcolor{red}{$1.2/0.48$} \\
    Kitchen2 & $1.2/0.5$ & $-$ & $1.55/0.70$ & $4.90/2.30$ & \textcolor{blue}{$1.30/0.55$} & \textcolor{red}{$0.81/0.35$} \\
    Living2 & $2.0/0.8$ & $-$ & $2.60/1.00$ & $23.7/9.53$ & \textcolor{blue}{$2.26/1.06$} & \textcolor{red}{$1.01/0.41$} \\
    Luke & $2.6/1.0$ & $-$ & $2.80/1.11$ & $232/95.05$ & \textcolor{blue}{$2.52/0.97$} & \textcolor{red}{$1.53/0.60$} \\
    Gates362 & $2.0/0.8$ & $-$ & $1.73/0.62$ & $180/69.70$ & \textcolor{blue}{$1.47/0.44$} & \textcolor{red}{$0.95/0.28$} \\
    Gates381 & $2.7/1.2$ & $-$ & \textcolor{blue}{$1.78/0.81$} & $290/103.7$ & $2.18/0.93$ & \textcolor{red}{$1.12/0.46$} \\
    Lounge & $1.8/0.6$ & $-$ & $2.29/0.77$ & $129/38.98$ & \textcolor{blue}{$2.00/0.70$} & \textcolor{red}{$1.11/0.35$} \\
    Manolis & $1.6/0.7$ & $-$ & $2.42/1.06$ & $215/80.19$ & \textcolor{blue}{$1.54/0.68$} & \textcolor{red}{$1.05/0.42$} \\
    Office5a & $2.5/0.9$ & $-$ & \textcolor{blue}{$2.40/0.94$} & $209/78.16$ & $2.89/1.29$ & \textcolor{red}{$1.96/0.89$} \\
    Office5b & $2.6/0.8$ & $-$ & $2.73/0.86$ & $9.30/2.87$ & \textcolor{blue}{$1.74/0.54$} & \textcolor{red}{$1.17/0.41$} \\
    \hline
  \end{tabular}
  \label{comparison2}
\end{table*}

\subsection{Localization Result}

In this section, we present localization results in scarce data scenarios, which are shown in detail in Table \ref{comparison} and \ref{comparison2}. Notably, except for the few-shot learning-based SRC method, all other approaches experience a decline in performance with a significant reduction in training samples. 

In Table \ref{comparison}, SCR-Net \cite{li_hierarchical_2020} yields the poorest results among the learning-based methods in Table \ref{comparison}, while HSC-Net \cite{li_hierarchical_2020} performs well, especially in the Chess scene, surpassing the few-shot approach. Combining D2S with our descriptor synthesis pipeline yields the best localization results for 4 out of 7 scenes. Compared to a similar NeRF method \cite{chen_leveraging_2023} with 25\% of training data, our approach uses less data and outperforms it in 3 out of 7 scenes, with comparable performance in others. However, in scenes with repeated patterns like Stairs, our method suffers and falls behind the few-shot approach due to descriptor similarity. 

The results in Table \ref{comparison2} for the 12Scenes dataset emphasize that our approach achieves state-of-the-art results across all scenes compared to other SCR approaches. In scenes like Bed, Gates381, and Office5a, where HSC-Net \cite{li_hierarchical_2020} outperforms the original D2S, our pipeline significantly aids D2S in achieving lower errors than HSC-Net. In comparison, our method achieves translation and rotation errors of 1.14 cm and 0.45$^{\circ}$, respectively, approximately half that of HSC-Net (which has errors of 2.18 cm and 0.87$^{\circ}$, respectively). Notably, our pipeline even outperforms \cite{chen_leveraging_2023} in every scene, despite the latter being trained with more data.

Comparing the original implementation of D2S \cite{bui_d2s_2023} (without synthetic data) to our method in the last two columns of Tables \ref{comparison} and \ref{comparison2}, improvements are observed across all scenes in both datasets. For 7Scenes, smaller volume scenes like Fire and Chess show a reduction in translation error of 20\% and 16\%, along with a rotation error reduction of 22\% and 6\%, respectively. Larger scenes pose more challenges, with translation error improvements of 7.5\% for Office and 8.8\% for Redkitchen. Notably, in the challenging Stairs scene, where D2S faces difficulties, our pipeline reduces the localization error from 24.19cm/5.23$^{\circ}$ to 19.75cm/4.85$^{\circ}$, accounting for 18.3\% and 7\% reduction, respectively. Overall, our pipeline brings an average improvement of about 15\% for translation errors and 13\% for rotation errors across all scenes.

In the 12Scenes results presented in Table \ref{comparison2}, the better image quality and the utilization of more images lead to remarkable improvements compared to the original D2S implementation. Living2 scene experienced a significant drop in localization error from 2.26cm/1.06$^{\circ}$ to 1.01cm/0.41$^{\circ}$, which is 55\% and 61\% improvement. The average improvement percentage for translation errors is 41.52\%, and for rotation errors, it is 42.8\%, surpassing the improvements observed in the 7Scenes dataset.

\section{CONCLUSIONS}

In this paper, we have introduced a data synthesis pipeline tailored for keypoint scene coordinate regressors. Leveraging advancements in the Neural Radiance Field, feature extraction, and feature matching, we've managed to enhance data generation from a limited set of samples. The combination of Nerfacto and LightGlue matcher offers speed, high-quality view synthesis, and effective view evaluation for the pipeline. D2S, enhanced with the additionally synthesized data, outperforms the few-shot and previous learning-based methods and showcases significant improvements compared to the original approach for both datasets in experiments. 

Despite achieving high accuracy, our method encounters challenges associated with NeRF. Outdoor environments present a formidable challenge due to dynamic factors like seasonal changes, variations in illumination, and moving objects. However, given the rapid evolution of Neural Rendering research and the flexibility of our method, newer models addressing these issues could seamlessly integrate into our pipeline, offering avenues for further improvement.





\bibliographystyle{IEEEtran}
\bibliography{HoangThuanIROS2024.bib}

\begin{thebibliography}{10}
\providecommand{\url}[1]{#1}
\csname url@rmstyle\endcsname
\providecommand{\newblock}{\relax}
\providecommand{\bibinfo}[2]{#2}
\providecommand\BIBentrySTDinterwordspacing{\spaceskip=0pt\relax}
\providecommand\BIBentryALTinterwordstretchfactor{4}
\providecommand\BIBentryALTinterwordspacing{\spaceskip=\fontdimen2\font plus
\BIBentryALTinterwordstretchfactor\fontdimen3\font minus \fontdimen4\font\relax}
\providecommand\BIBforeignlanguage[2]{{%
\expandafter\ifx\csname l@#1\endcsname\relax
\typeout{** WARNING: IEEEtran.bst: No hyphenation pattern has been}%
\typeout{** loaded for the language `#1'. Using the pattern for}%
\typeout{** the default language instead.}%
\else
\language=\csname l@#1\endcsname
\fi
#2}}

\bibitem{schonberger2016structure}
J.~L. Schonberger and J.-M. Frahm, ``Structure-from-motion revisited,'' in \emph{Proceedings of the IEEE conference on computer vision and pattern recognition}, 2016, pp. 4104--4113.

\bibitem{bui_d2s_2023}
B.-T. Bui, D.-T. Tran, and J.-H. Lee, ``\BIBforeignlanguage{en}{{D2S}: {Representing} local descriptors and global scene coordinates for camera relocalization},'' Dec. 2023, arXiv:2307.15250 [cs].

\bibitem{kendall_posenet_2016}
A.~Kendall, M.~Grimes, and R.~Cipolla, ``Posenet: A convolutional network for real-time 6-dof camera relocalization,'' in \emph{Proceedings of the IEEE international conference on computer vision}, 2015, pp. 2938--2946.

\bibitem{bach_featloc_2022}
T.~B. Bach, T.~T. Dinh, and J.-H. Lee, ``\BIBforeignlanguage{en}{{FeatLoc}: {Absolute} pose regressor for indoor {2D} sparse features with simplistic view synthesizing},'' \emph{\BIBforeignlanguage{en}{ISPRS Journal of Photogrammetry and Remote Sensing}}, vol. 189, pp. 50--62, July 2022.

\bibitem{zhou_kfnet_2020}
L.~Zhou, Z.~Luo, T.~Shen, J.~Zhang, M.~Zhen, Y.~Yao, T.~Fang, and L.~Quan, ``Kfnet: Learning temporal camera relocalization using kalman filtering,'' in \emph{Proceedings of the IEEE/CVF conference on computer vision and pattern recognition}, 2020, pp. 4919--4928.

\bibitem{zhou2020learn}
Q.~Zhou, T.~Sattler, M.~Pollefeys, and L.~Leal-Taixe, ``To learn or not to learn: Visual localization from essential matrices,'' in \emph{2020 IEEE International Conference on Robotics and Automation (ICRA)}.\hskip 1em plus 0.5em minus 0.4em\relax IEEE, 2020, pp. 3319--3326.

\bibitem{brachmann_dsac_2018}
E.~Brachmann, A.~Krull, S.~Nowozin, J.~Shotton, F.~Michel, S.~Gumhold, and C.~Rother, ``Dsac-differentiable ransac for camera localization,'' in \emph{Proceedings of the IEEE conference on computer vision and pattern recognition}, 2017, pp. 6684--6692.

\bibitem{brachmann_learning_2018}
E.~Brachmann and C.~Rother, ``Learning less is more-6d camera localization via 3d surface regression,'' in \emph{Proceedings of the IEEE conference on computer vision and pattern recognition}, 2018, pp. 4654--4662.

\bibitem{li_hierarchical_2020}
X.~Li, S.~Wang, Y.~Zhao, J.~Verbeek, and J.~Kannala, ``Hierarchical scene coordinate classification and regression for visual localization,'' in \emph{Proceedings of the IEEE/CVF Conference on Computer Vision and Pattern Recognition}, 2020, pp. 11\,983--11\,992.

\bibitem{dong_visual_2022}
S.~Dong, S.~Wang, Y.~Zhuang, J.~Kannala, M.~Pollefeys, and B.~Chen, ``Visual localization via few-shot scene region classification,'' in \emph{2022 International Conference on 3D Vision (3DV)}.\hskip 1em plus 0.5em minus 0.4em\relax IEEE, 2022, pp. 393--402.

\bibitem{kukelova2013real}
Z.~Kukelova, M.~Bujnak, and T.~Pajdla, ``Real-time solution to the absolute pose problem with unknown radial distortion and focal length,'' in \emph{Proceedings of the IEEE International Conference on Computer Vision}, 2013, pp. 2816--2823.

\bibitem{sarlin_coarse_2019}
P.-E. Sarlin, C.~Cadena, R.~Siegwart, and M.~Dymczyk, ``From coarse to fine: Robust hierarchical localization at large scale,'' in \emph{Proceedings of the IEEE/CVF Conference on Computer Vision and Pattern Recognition}, 2019, pp. 12\,716--12\,725.

\bibitem{mildenhall_nerf_2020}
B.~Mildenhall, P.~P. Srinivasan, M.~Tancik, J.~T. Barron, R.~Ramamoorthi, and R.~Ng, ``Nerf: Representing scenes as neural radiance fields for view synthesis,'' \emph{Communications of the ACM}, vol.~65, no.~1, pp. 99--106, 2021.

\bibitem{moreau_lens_nodate}
A.~Moreau, N.~Piasco, D.~Tsishkou, B.~Stanciulescu, and A.~de~La~Fortelle, ``Lens: Localization enhanced by nerf synthesis,'' in \emph{Conference on Robot Learning}.\hskip 1em plus 0.5em minus 0.4em\relax PMLR, 2022, pp. 1347--1356.

\bibitem{tancik_nerfstudio_2023}
M.~Tancik, E.~Weber, E.~Ng, R.~Li, B.~Yi, J.~Kerr, T.~Wang, A.~Kristoffersen, J.~Austin, K.~Salahi, A.~Ahuja, D.~McAllister, and A.~Kanazawa, ``\BIBforeignlanguage{en}{Nerfstudio: {A} {Modular} {Framework} for {Neural} {Radiance} {Field} {Development}},'' in \emph{\BIBforeignlanguage{en}{Special {Interest} {Group} on {Computer} {Graphics} and {Interactive} {Techniques} {Conference} {Conference} {Proceedings}}}, July 2023, pp. 1--12, arXiv:2302.04264 [cs].

\bibitem{detone_superpoint_2018}
D.~DeTone, T.~Malisiewicz, and A.~Rabinovich, ``\BIBforeignlanguage{en}{{SuperPoint}: {Self}-{Supervised} {Interest} {Point} {Detection} and {Description}},'' Apr. 2018, arXiv:1712.07629 [cs].

\bibitem{lindenberger_lightglue_2023}
P.~Lindenberger, P.-E. Sarlin, and M.~Pollefeys, ``Lightglue: Local feature matching at light speed,'' \emph{arXiv preprint arXiv:2306.13643}, 2023.

\bibitem{schonberger2016pixelwise}
J.~L. Sch{\"o}nberger, E.~Zheng, J.-M. Frahm, and M.~Pollefeys, ``Pixelwise view selection for unstructured multi-view stereo,'' in \emph{Computer Vision--ECCV 2016: 14th European Conference, Amsterdam, The Netherlands, October 11-14, 2016, Proceedings, Part III 14}.\hskip 1em plus 0.5em minus 0.4em\relax Springer, 2016, pp. 501--518.

\bibitem{tyszkiewicz_disk_2020}
M.~Tyszkiewicz, P.~Fua, and E.~Trulls, ``Disk: Learning local features with policy gradient,'' \emph{Advances in Neural Information Processing Systems}, vol.~33, pp. 14\,254--14\,265, 2020.

\bibitem{dusmanu_d2-net_2019}
M.~Dusmanu, I.~Rocco, T.~Pajdla, M.~Pollefeys, J.~Sivic, A.~Torii, and T.~Sattler, ``D2-net: A trainable cnn for joint detection and description of local features,'' \emph{arXiv preprint arXiv:1905.03561}, 2019.

\bibitem{revaud_r2d2_2019}
J.~Revaud, P.~Weinzaepfel, C.~De~Souza, N.~Pion, G.~Csurka, Y.~Cabon, and M.~Humenberger, ``R2d2: repeatable and reliable detector and descriptor,'' \emph{arXiv preprint arXiv:1906.06195}, 2019.

\bibitem{arandjelovic_netvlad_2016}
R.~Arandjelovic, P.~Gronat, A.~Torii, T.~Pajdla, and J.~Sivic, ``Netvlad: Cnn architecture for weakly supervised place recognition,'' in \emph{Proceedings of the IEEE conference on computer vision and pattern recognition}, 2016, pp. 5297--5307.

\bibitem{gordo_end--end_2017}
A.~Gordo, J.~Almazan, J.~Revaud, and D.~Larlus, ``End-to-end learning of deep visual representations for image retrieval,'' \emph{International Journal of Computer Vision}, vol. 124, no.~2, pp. 237--254, 2017.

\bibitem{sarlin_superglue_2020}
P.-E. Sarlin, D.~DeTone, T.~Malisiewicz, and A.~Rabinovich, ``Superglue: Learning feature matching with graph neural networks,'' in \emph{Proceedings of the IEEE/CVF conference on computer vision and pattern recognition}, 2020, pp. 4938--4947.

\bibitem{bergamo2013leveraging}
A.~Bergamo, S.~N. Sinha, and L.~Torresani, ``Leveraging structure from motion to learn discriminative codebooks for scalable landmark classification,'' in \emph{Proceedings of the IEEE Conference on Computer Vision and Pattern Recognition}, 2013, pp. 763--770.

\bibitem{brahmbhatt_geometry-aware_2018}
S.~Brahmbhatt, J.~Gu, K.~Kim, J.~Hays, and J.~Kautz, ``Geometry-aware learning of maps for camera localization,'' in \emph{Proceedings of the IEEE conference on computer vision and pattern recognition}, 2018, pp. 2616--2625.

\bibitem{wang2020atloc}
B.~Wang, C.~Chen, C.~X. Lu, P.~Zhao, N.~Trigoni, and A.~Markham, ``Atloc: Attention guided camera localization,'' in \emph{Proceedings of the AAAI Conference on Artificial Intelligence}, vol.~34, no.~06, 2020, pp. 10\,393--10\,401.

\bibitem{sattler_understanding_2019}
T.~Sattler, Q.~Zhou, M.~Pollefeys, and L.~Leal-Taixe, ``Understanding the limitations of cnn-based absolute camera pose regression,'' in \emph{Proceedings of the IEEE/CVF conference on computer vision and pattern recognition}, 2019, pp. 3302--3312.

\bibitem{ng_reassessing_2021}
T.~Ng, A.~Lopez-Rodriguez, V.~Balntas, and K.~Mikolajczyk, ``Reassessing the limitations of cnn methods for camera pose regression,'' \emph{arXiv preprint arXiv:2108.07260}, 2021.

\bibitem{brachmann2021dsacstar}
E.~Brachmann and C.~Rother, ``Visual camera re-localization from {RGB} and {RGB-D} images using {DSAC},'' \emph{TPAMI}, 2021.

\bibitem{pittaluga_revealing_2019}
F.~Pittaluga, S.~J. Koppal, S.~B. Kang, and S.~N. Sinha, ``\BIBforeignlanguage{en}{Revealing {Scenes} by {Inverting} {Structure} from {Motion} {Reconstructions}},'' Apr. 2019, arXiv:1904.03303 [cs].

\bibitem{zhang2022rendernet}
J.~Zhang, S.~Tang, K.~Qiu, R.~Huang, C.~Fang, L.~Cui, Z.~Dong, S.~Zhu, and P.~Tan, ``Rendernet: Visual relocalization using virtual viewpoints in large-scale indoor environments,'' \emph{arXiv preprint arXiv:2207.12579}, 2022.

\bibitem{liu_posegan_2020}
K.~Liu, Q.~Li, and G.~Qiu, ``Posegan: A pose-to-image translation framework for camera localization,'' \emph{ISPRS Journal of Photogrammetry and Remote Sensing}, vol. 166, pp. 308--315, 2020.

\bibitem{chen_leveraging_2023}
L.~Chen, W.~Chen, R.~Wang, and M.~Pollefeys, ``Leveraging neural radiance fields for uncertainty-aware visual localization,'' \emph{arXiv preprint arXiv:2310.06984}, 2023.

\bibitem{barron_mip-nerf_2021}
J.~T. Barron, B.~Mildenhall, M.~Tancik, P.~Hedman, R.~Martin-Brualla, and P.~P. Srinivasan, ``\BIBforeignlanguage{en}{Mip-{NeRF}: {A} {Multiscale} {Representation} for {Anti}-{Aliasing} {Neural} {Radiance} {Fields}},'' in \emph{\BIBforeignlanguage{en}{2021 {IEEE}/{CVF} {International} {Conference} on {Computer} {Vision} ({ICCV})}}.\hskip 1em plus 0.5em minus 0.4em\relax Montreal, QC, Canada: IEEE, Oct. 2021, pp. 5835--5844.

\bibitem{barron_mip-nerf_2022}
J.~T. Barron, B.~Mildenhall, D.~Verbin, P.~P. Srinivasan, and P.~Hedman, ``Mip-nerf 360: Unbounded anti-aliased neural radiance fields,'' in \emph{Proceedings of the IEEE/CVF Conference on Computer Vision and Pattern Recognition}, 2022, pp. 5470--5479.

\bibitem{muller_instant_2022}
T.~Müller, A.~Evans, C.~Schied, and A.~Keller, ``\BIBforeignlanguage{en}{Instant {Neural} {Graphics} {Primitives} with a {Multiresolution} {Hash} {Encoding}},'' \emph{\BIBforeignlanguage{en}{ACM Transactions on Graphics}}, vol.~41, no.~4, pp. 1--15, July 2022, arXiv:2201.05989 [cs].

\bibitem{wang2021nerf}
Z.~Wang, S.~Wu, W.~Xie, M.~Chen, and V.~A. Prisacariu, ``Nerf--: Neural radiance fields without known camera parameters,'' \emph{arXiv preprint arXiv:2102.07064}, 2021.

\bibitem{shotton_scene_2013}
J.~Shotton, B.~Glocker, C.~Zach, S.~Izadi, A.~Criminisi, and A.~Fitzgibbon, ``Scene coordinate regression forests for camera relocalization in rgb-d images,'' in \emph{Proceedings of the IEEE conference on computer vision and pattern recognition}, 2013, pp. 2930--2937.

\bibitem{valentin2016learning}
J.~Valentin, A.~Dai, M.~Nie{\ss}ner, P.~Kohli, P.~Torr, S.~Izadi, and C.~Keskin, ``Learning to navigate the energy landscape,'' in \emph{2016 Fourth International Conference on 3D Vision (3DV)}.\hskip 1em plus 0.5em minus 0.4em\relax IEEE, 2016, pp. 323--332.

\end{thebibliography}

\end{document}